**Authors:**
Sultan Sevgi Turgut[1*], M.Sc. and
Mustafa Oral[2], Ph.D.

**Title:**
Multi-Focus Image Fusion based on Gradient Transform

**Affiliations:**
[1]Yildiz Technical University, Computer Engineering Department, Istanbul, Turkey
[2] Cukurova University, Computer Engineering Department, Adana, Turkey

**Correspondence:**
[*]Sultan Sevgi Turgut, ORCID ID: 0000-0002-1422-9186, E-mail: sturgut@yildiz.edu.tr

# Multi-Focus Image Fusion based on Gradient Transform

Sultan Sevgi Turgut[1], Mustafa Oral[2]

[1]Department of Computer Engineering, Yildiz Technical University, Istanbul, Turkey

[2]Department of Computer Engineering, Cukurova University, Adana, Turkey

**Abstract**

Multi-focus image fusion is a challenging field of study that aims to provide a completely focused image by integrating focused and un-focused pixels. Most existing methods suffer from shift variance, misregistered images, and data-dependent. In this study, we introduce a novel gradient information-based multi-focus image fusion method that is robust for the aforementioned problems. The proposed method first generates gradient images from original images by using Halftoning-Inverse Halftoning (H-IH) transform. Then, Energy of Gradient (EOG) and Standard Deviation functions are used as the focus measurement on the gradient images to form a fused image. Finally, in order to enhance the fused image a decision fusion approach is applied with the majority voting method. The proposed method is compared with 17 different novel and conventional techniques both visually and objectively. For objective evaluation, 6 different quantitative metrics are used. It is observed that the proposed method is promising according to visual evaluation and 83.3% success is achieved by being first in five out of six metrics according to objective evaluation.

**Keywords**

*Image processing; multi-focus image fusion; the energy of gradient; halftoning.*

## 1 INTRODUCTION

Image fusion is an important branch of image processing. The fusion field has four different types; multi-view(mVF), multi-modal(mMF), multi-temporal(mTF) and multi-focus(mFF). mVF fuses different views of the same scene while mMF fuses images from different modalities to reveal principal information. mTF allows observing alternation of a scene over time. mFF fuses images with different focal lengths to produce a fully focused image [1]. This study concentrates on the mFF.

Due to the limited depth of field feature of the optical lenses, it allows focusing on only one region at a time. Therefore, it is quite challenging to get fully focused images. For this reason, images consist of regions with different focal levels. Several methods have been developed to solve this problem [2].

Fusion methods can generally be constructed on three levels; pixel level, feature level, and decision level. The pixel-level methods directly apply the fusion operation without any processing on the source images. This level includes simple approximation methods. Feature level that is the intermediate complexity level extracts features such as illumination, edges, directions etc. from the images instead of fusing the source images. The fusion step is accomplished with these features. The purpose of feature extraction is to clean the images from unnecessary information and compare them at a common point. The highest level of complexity decision level extracts features from images and transmits to the feature identification process. In this process, the features are processed with the help of various methods, and then the fusion process is applied [3].

Multi-Focus Image Fusion Algorithms are divided into two domains; spatial domain and transformation domain methods. Spatial domain methods implement the fusion process without transforming images into another domain. These methods are more resistant to effects such as registration disorder and noise. On the other hand, these methods can cause spectral distortion and spatial degradation [4]. In the transformation domain, before the image fusion, the source images are transformed into a different scale domain. In some cases, transformation is applied to process the image in multi-scale. Transform domain-based methods are more sensitive to registration errors, and these methods are more time-consuming due to the transformation complexity [5].

Morris and Rajesh have proposed three well-known spatial domain methods; Averaging, Min/Max selection and Principal Component Analysis (PCA) [6]. In Averaging method, averaging pixels of source images is used to create fused image, in Min/Max methods, pixels with minimum or maximum intensities are selected to achieve same goal. The main use of PCA is to reduce high dimensional data with correlated variables into a low dimension with more variant structure without loss of information. Unique Color (UC) [7]. and Standard Deviation Method (Std) [8] are good examples of successful spatial algorithms. These two methods have a similar perspective. Both methods are based on the principle of diversity as a method of detecting focused pixels.

Naidu et al. have proposed important transform domain methods, Discrete Cosine Transform (DCT) and Discrete Wavelet Transform (DWT) [9]-[10]. These methods transform data from spatial domain to frequency domain. Pixels' frequencies is useful and important where distinguishing details and outlines in the images. DCT uses only cosine component for decomposition so it has a cost-effective approach. On the other hand, DWT can diminish spatial distortion. Zhang et al. have developed Nonsubsampled Contourlet Transform (NSCT) as a different type of transform domain method [11]. They have focused two properties on their algorithms; directionality and anisotropy. NSCT is a shift-invariance method and so it can be robust for misregistered images. Through with its directionality, it can represent images with much less coefficient according to DWT and DCT. However, it is a nonpersistent method that is not robust to even by minor changes. Another example of conventional transform domain algorithms is pyramid-based ones; Laplacian Pyramid [12], Morphological Pyramid [13], Ratio Pyramid [14]. These methods aim to achieve details of images by means of lowpass and bandpass filtering.

Liu et al. are researchers who developed the fusion algorithm benefit from deep learning methods [15]. They aim to perform both feature extraction and classification operations together using Convolutional Neural Network (CNN). They have inspired other studies in the literature with their successful results. However, the requirement for large data sets and the training process are the disadvantages of their method. An example of neural network model, Pulse Coupled Neural Network [16] has attracted a lot of attention over time and has been developed by various researchers; Dual-Channel PCNN [17], Multi-Channel PCNN [18]. The PCNN model based fusion method [19] is combined with existing fusion methods to achieve successful results. One of them is the combining of PCNN and Laplacian Pyramid (LP-PCNN) [20]. However, researchers could not avoid the performance burden of combining more than one method.

According to the literature, it is stated that while details are prominent in focused areas, it is difficult to capture the details in unfocused areas. Therefore, methods utilized gradient information have published in recent years. He et al. have developed a new approach to transformation methods with the Guided Filter Method [21]. They have made multi-scale transformations by filtering instead of transform operations. Guided Filter serves to reveal gradient information and eliminate noise in the image by variance information. Liu et al. contribute to the literature with the Convolutional Sparse Representation that performs transformation by filtering [22]. This method has been developed in order to obtain details more successfully and find solutions to misregistered images. Another example of filtering methods is the Cross Bilateral Filtering that is proposed by Kumar [23]. The difference of CBF is that it uses both similarities in color space and proximity in geometric space. Bavirisetti and Dhuli aim to detect saliency information and thus get rid of noises by using the average filter in Multi-Scale Image Decomposition and Saliency Detection Method (MSSF) [24]. It provides performance gain with its simple filtering structure. One of the outstanding studies by processing gradient information has been developed by Zhou and Wang [25]. Multi-Scale Weighted Gradient-Based Method (MWGF) concentrates on anisotropic blur and misregistration problems.

In this study, our objective is to examine present mFF techniques and carry out comparative work to demonstrate the success rates of the methods. It is also aimed to propose a novel method with a new perspective taking advantage of the existing methods. The proposed method is shift-invariant, resistant to misregistered images, data-independent, and can capture details in focus. Our method consists of three steps; preprocess, initial fusion, and decision-level fusion. Preprocess highlights important details by eliminating the noise in the image. We concentrate on two features contrast and gradient information. As a result of preprocess, a gradient image is obtained by applying histogram equalization and Halftoning-Inverse Halftoning (H-IH)transform. In the initial fusion step, a focus measurement method on the gradient image is applied with a sliding window. The Energy of Gradient (EOG) and Standard Deviation functions are implemented. The initial fused image is obtained by fusing blocks with high measurement values of source images. In order to improvement of initial fusion, the decision fusion step also known as consistency verification is applied. In decision fusion mechanism, majority voting system, voting is performed by implementing different block sizes, is implemented. As a result, a fully focused image is obtained.

The paper is designed as firstly an overview is presented to the summaries methods section, then methods and quantitative performance metrics are explained in detail. Finally, visual and objective results are shown and discussed.

## 2 METHODOLOGY AND PROPOSED FUSION SCHEME

### 2.1 Overview

In this study, we focus on two features that are highly emphasized in the literature and our experiments; image contrast and gradient information. Contrast enhancement as a pre-processing operation has been applied to the source images to make the image details more prominent. To accomplish this, we use histogram equalization method. Histogram equalization ensures regular distribution of image information and makes the image more convenient to capture details.

Another important feature; gradient information. Unnecessary information should be discarded and the remaining information, the details, treated as a summary of the original source image. There are numerous edge detectors such as Canny, Sobel, Laplacian, LOG, etc. The fusion algorithms that use edge information, employ one of these traditional edge detectors or filters to extract edge information. Contrary to the current literature, we aim to concentrate on gradient rather than edge information, so we propose to employ an unconventional method to extract gradient information. The gradient information is conventionally extracted from local information preserved in an image. However, we get the gradient information from a pair of images that are the source image and its counterpart that is obtained by applying Halftoning-Inverse Halftoning (H-IH)transform consecutively.

Detection of focused and non-focused pixels is achieved by applying Energy of Gradient (EOG) as a focus measure function. Higher values of gradient energy are the indicator of in focus pixels. However, in some cases, the result of the focus measurement of the compared pixel blocks may be equal. In such cases, the second focus measure method is used to avoid losing information. Standard deviation is selected as the second focus measure.

The success of the proposed method can be enhanced with decision-level fusion mechanism. Many practical studies in the literature implement a consistenciy verification step. In decision-level fusion, we use majority voting method. Figure 1 illustrates the steps of the proposed method.

### 2.2 Local Histogram Equalization

Image enhancement is a crucial operation for image processing applications. Improving image information empowers the appearance of key features of the image, simplifies operations, and makes results more effective. One of the image enhancement methods is histogram equalization. The histogram is a graphical representation of the distribution of gray level in the image. Equalization means spreading the most frequent pixel values and extending the intensity range of the image. Thus, the contrast value of details is increased by means of a histogram equalization technique. Histogram equalization is generally examined into two ways; Global (GHE) and Local Histogram Equalization (LHE). In GHE, all image information is processed together. Thus, it provides an overall contrast enhancement so it causes loss of characteristic local information such as brightness. In order to find a solution to this problem, LHE technique has been developed. The difference of the LHE method is that the image is processed based on the blocks [26]-[28].

The sequence of histogram equalization operations is as follows; First of all, the histogram of the image is created. Then, the Cumulative Density Function (CDF) is calculated. The cumulative density is the magnitude that includes the values obtained by the sum of the previous and each value of the histogram itself. By using these cumulative density values, new gray level values are achieved as in Equations (1 and 2).

$$p_n = \frac{number\ of\ pixels\ in\ the\ n^{th}\ gray\ level}{total\ number\ of\ pixels} \quad (1)$$

$$g_k = floor\left((L-1)\sum_{n=0}^{k} p_n\right), \ k = 0,1,2,\ldots,L-1 \tag{2}$$

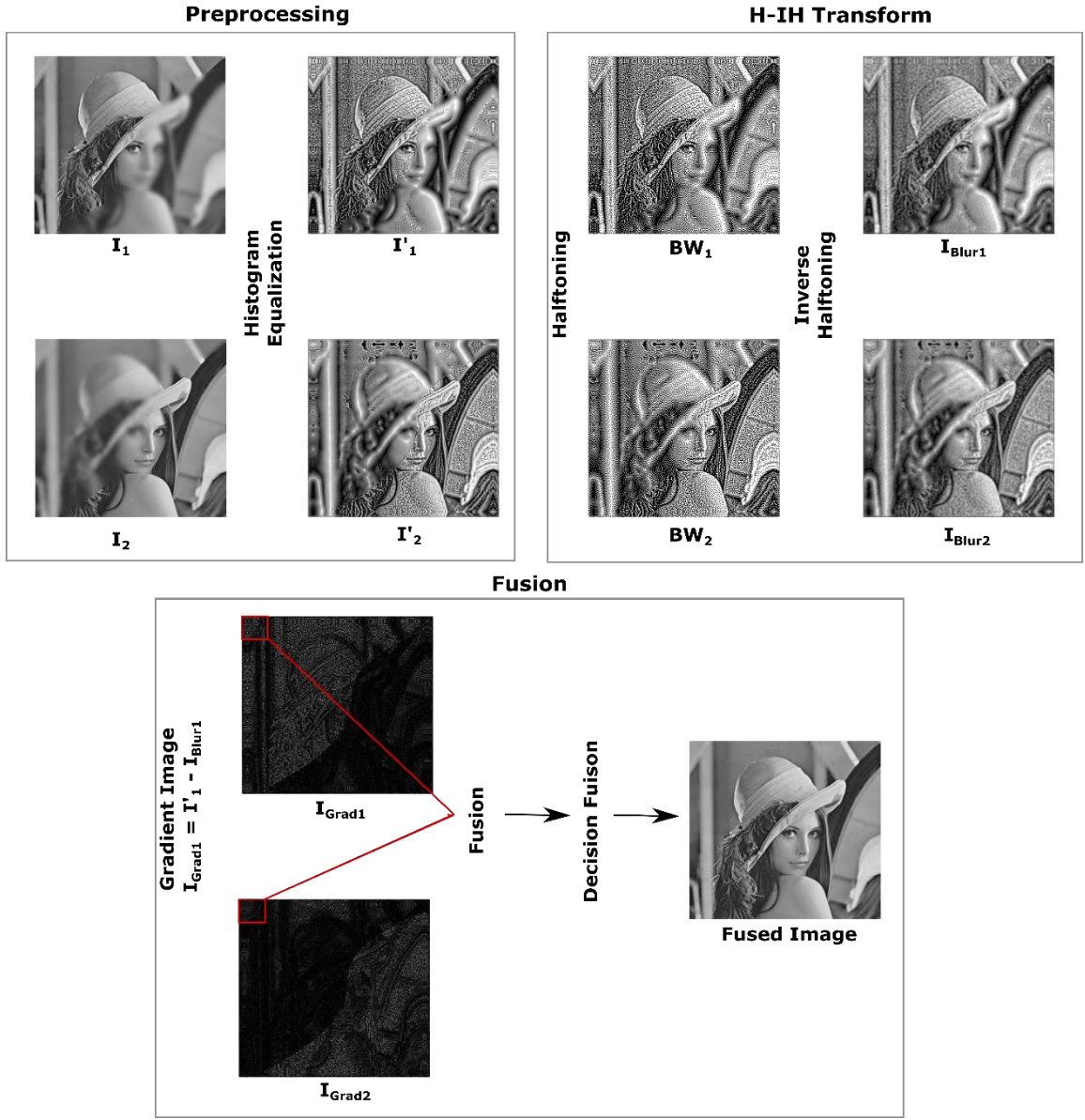

Figure 1 - Overview of the proposed method.

In the fusion process, due to the disadvantages of global histogram equalization, we preferred local histogram equalization. $I$ is the original source image, and $I'$ is the version of applied local histogram equalization.

$$I' = LHE(I) \tag{3}$$

## 2.3 Halftoning

Digital Image Halftoning or Dithering [29] provides converting a continuous-tone image to a binary-valued (black and white) image. This technique places the black microdots in appropriate regions to generate a binary image that is visually similar to the continuous-tone image. Due to the low pass frequency property, the human eyes perceive the grey-toned patches, instead of the microdots. Thus, halftoning create an optical illusion, and a small size image that has similar visual to the original image is obtained. So it may be considered as image compression techniques. However, halftoning techniques are generally used in printing, facsimile (FAX), scanner, computer graphics, indifferent image processing and electronic applications [30].

One of the most popular halftoning techniques is the error diffusion method. This study field dates back to 1976. Error diffusion method tries to reduce the total error and the image is reshaped with pure black(0) dots. The working principle of the algorithm is as follows; operations start to top-left of the image. First of all, the new value of the pixel is decided to be black or white as in Equation (4). In the equation, 0 < $p_{i,j}$ ≤ M refers to the related pixel of the continuous image and M is the maximum value that the pixel can take. $p'_{i,j}$ is the new value of the pixel. Then the error is calculated based on the new value of the original as in Equation (5). And this error is distributed to the neighbors of the pixel, Equation (6). The error is only distributed to unprocessed neighbors. Finally, the alpha $α_{x,y}$ parameter is the diffusion matrix coefficient corresponding to the neighbor pixel.

$$p'_{i,j} = \begin{cases} 1, p_{i,j} \geq M/2 \\ 0, otherwise \end{cases} \quad (4)$$

$$E_{i,j} = p_{i,j} - p'_{i,j} \quad (5)$$

$$p'_{i+x,j+y} = p_{i+x,j+y} + E_{i,j} * α_k \quad (6)$$

Error Diffusion has been investigated by several researchers, especially Floyd [31], Ninke [32], and Stucki et al [33]. We used the Floyd-Steinberg Error Diffusion Method, which is an effective and highly preferred method. In Equation 7, BW is the halftoned image of the $I'$ that applied local histogram equalizaton.

$$BW = Halftoning(I') \quad (7)$$

## 2.4 Inverse Halftoning

Halftoned images are not proper to be used in many image processing applications. The inverse halftoning methods convert the binary image to continuous form. The major problem of inverse halftoning, there can be more than one grayscale version of a halftoned image. Even if the applied halftoning algorithm is known, a single fixed continuous image cannot be obtained. Another problem is that inverse halftoning does not attract as much attention as halftoning among researchers. In the literature, the original images are commonly considered lowpass. However, applying a lowpass filter to halftoned images, we can lose the important detail information at the same time. Therefore, the inverse halftoned process is not very easy, and the multiple features of the image need to be examined [34]. Inverse Halftoning methods are divided into two categories; Dithering Model-Based Inverse (DMI) and Not based on the Dithering Model Inverse (NDMI). The methods in the DMI category implement inverse halftoning according to the applied halftoning algorithm. These algorithms have pre-knowledge and produce solutions depend on this. NDMI algorithms process as filtering methods or removing dithering noise. Since these algorithms develop inverse halftoning methods without prior knowledge, their performance is restricted.

We proceed with the inverse halftoning technique to access the gradient information. The inverse halftoning technique can be considered as filtering process. However, a suitable kernel must be created. Here, instead of theoretically observations, we can focus on the behaviour of the real human eye. We use the Human Vision System (HVS) [35]. There are various HVS models. One of the most similar models of the human visual system is the Gaussian HVS model. Gaussian HVS is described by spatial frequency response, also called a Modulation Transfer Function (MTF). Its formula is given in Equation (8). The inversed image is generated from the convolution of the dithered image by the HVS filter.

The process can be summarized as follows; firstly, the dithered image (BW) is generated by Floyd-Steinberg Error Diffusion. Then the inverse halftoning method is applied and an inversed image (IBlur) that is outlines of the original image is achieved as in Equation (9). Inverse Halftoning example is given in Figure 2. Finally, when the difference between the image (I') and inverse halftoned (IBlur) image is taken, the gradient image (IGrad) is calculated as in Equation (10), and presented in Figure 3.

$$h = e^{-(x^2+y^2)/2σ^2} \quad (8)$$

$$IBlur = BW * h \quad (9)$$

$$IGrad = I' - IBlur \quad (10)$$

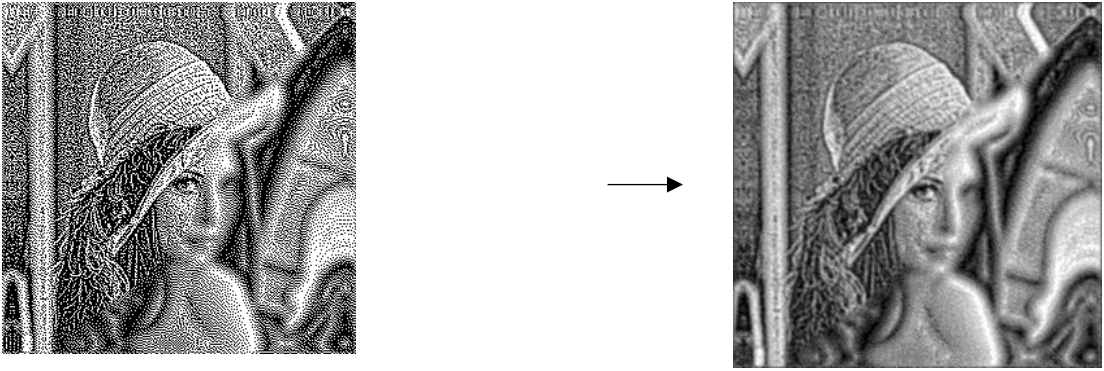

Figure 2 - Illustration of inverse halftoning

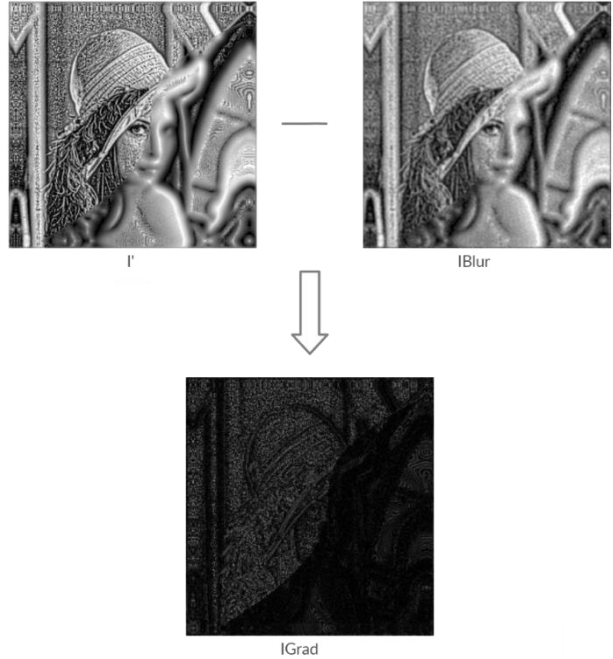

Figure 3 - Generating of gradient image.

## 2.5 Focus Measure

The key point of a fusion method is the focus measure. An appropriate focus measure is used to distinguish focused image blocks from unfocused image blocks. Succesful focus measures have some rules as folows; invariability according to image structure, robust to noises and artifacts, unimodal (the measurement must have only one peak value), efficient computational time and complexity

We have concentrated on high-frequency information (gradients) of the images. Correspondingly, gradient-based focus measurements are mainly examined. We utilize the Energy of Gradient Function (EOG), its formula is given in Equation 11 [36]- [37].

$$F_{EG} = \sum_r \sum_c (I_r^2 + I_c^2) \qquad (11)$$

where $I_r$ and $I_c$ correspond to gradient information at r and c positions in the image.

Energy evaluates how harmonious the gradient and image data. It is specified as the measure of the gradient. The energy function is applied block by block. The EOG of the respective block for each source images is calculated, and the centre pixel of the block that has the most energy is directly placed in the fused image. The illustration is given in Figure 4. In the equation, i and j are coordinates of source images $I_1$ and $I_2$.

$$Fused(i,j) = \begin{cases} I_1(i,j), EOG(I_1) > EOG(I_2) \\ I_2(i,j), \quad\quad\quad otherwise \end{cases} \quad (12)$$

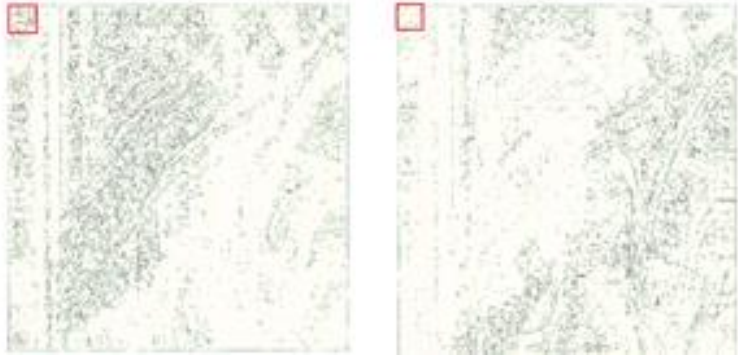

Figure 4 - Applying EOG Function block by block to the gradient versions of source images.

### 2.6 Second Focus Measure

The energy values in the compared blocks of the images may be the same. In such cases, an alternative focus measurement method is required. Blurred parts of images have a lower distribution of colors than the focused ones. Therefore, the standard deviation is a distinctive metric, it has been considered as an alternative focus measurement in this study. When the EOG function cannot decide the focus block, we use I' (applied local histogram image) with standard deviation function and the block which has the higher value is selected.

### 2.7 Decision-Level Fusion

Decision-level fusion, which is supported by studies in the literature, is added to the our method. It provides to enhances the fusion system. The result makes the fused image more accurate and effective.

Before making decision-level fusion, a fused map is generated from the indices that correspond to source images pixels. To accomplish decision making, we applied three different filters with 2x2, 4x4, 8x8 window sizes. Each filter has been used to apply a pooling operation. After all three filters have been applied, the majority voting [38] has been used to combine the results of filter processing.

## 3 PERFORMANCE METRICS

A procedure to prove the success of a method is objective evaluation. Various metrics are available for this purpose. The common feature of these metrics is that they do not need a reference picture. Metrics are examined in four categories; information theory-based, image structural similarity-based, human perception-based, image feature-based. In this study, six metrics are selected to make assessments from these categories. These;

i. Information Theory-Based:

Normalized Mutual Information ($Q_{NMI}$),

Nonlinear Correlation Information Entropy Metric ($Q_{NCIE}$).

ii. Image Feature Based:
Xyedas-Petrovic Metric ($Q_{AB/F}$).
Peng-Wei Wang Metric ($Q_{PWW}$).

iii. Image Structural Similarity Based:

Yang Metric ($Q_Y$).

iv. Human Perception Based:

Chen-Blum Metric ($Q_{CB}$)

### 3.1 Mutual Information (MI)

Mutual Information (MI) [39] determines the similarity relations between two discrete variables. This is described as Kullback–Leibler Divergence (relative entropy) that is the distance concerning the probability distribution of two variables in mathematics. In the image fusion, MI is used as a metric to gain performance information about algorithms.

In the fusion process, the amount of information obtained is taken into consideration. That is, the mutual information value is extracted between the fused image and the input images, and these values are summed to get the total information in the fused image. For performance comparison, the image including the highest MI value is considered more successful and knowledgeable. Kullback–Leibler Divergence is used for obtaining the joint distribution of fused and source images, then these values are added.

$$M_F^{AB} = I_{FA}(f,a) + I_{FB}(f,b) \tag{13}$$

### 3.2 Nonlinear Correlation Information Entropy Metric (NCIE)

A Nonlinear Correlation Coefficient (NCC) [40]-[41] defines the relationship between two discrete values of the multivariable data set by means of creating a nonlinear correlation matrix. NCC is similar to Mutual Information (MI) and it produces a number that is in the interval [0,1], bigger values indicates strong relationship.

$$R = \begin{pmatrix} NCC_{AA} & NCC_{AB} & NCC_{AF} \\ NCC_{BA} & NCC_{BB} & NCC_{BF} \\ NCC_{FA} & NCC_{FB} & NCC_{FF} \end{pmatrix} = \begin{pmatrix} 1 & NCC_{AB} & NCC_{AF} \\ NCC_{BA} & 1 & NCC_{BF} \\ NCC_{FA} & NCC_{FB} & 1 \end{pmatrix}$$

The NCIE of the R matrix is calculated as:

$$NCIE = 1 + \sum_{i=1}^{3} \frac{\lambda_i}{3} \log_b \frac{\lambda_i}{3} \tag{14}$$

Eigen values of the matrix R is represented by $\lambda_i$, (i= 1,2,3).

### 3.3 Xydeas and Petrovic Metric

Xydeas and Petrovic [42]-[43] have introduced a fusion metric that measures pixel-based performance. This metric calculates the amount of edge information transferred from the source images to the resulting image. So it performs visual comprehension just like the human eye. When performing on edge operation, it disregards regional information. Images that have more edge information transferred, in other words, high metric values, are considered successful. Sobel edge detection method is made use of to acquire the edge information.

In the image fusion, edge preservation values of two source images $Q^{AF}(i,j) = Q_g^{AF}(i,j) + Q_\alpha^{AF}(i,j)$ and $Q^{BF}(i,j) = Q_g^{BF}(i,j) + Q_\alpha^{BF}(i,j)$ with respect to the fused image are calculated and normalized with weights $w^A(i,j)$ and $w^B(i,j)$ as follows:

$$Q^{AB/F} = \frac{\sum_{i=1}^{N}\sum_{j=1}^{M} Q^{AF}(i,j)w^A(i,j) + Q^{BF}(i,j)w^B(i,j)}{\sum_{i=1}^{N}\sum_{j=1}^{M}(w^A(i,j) + w^B(i,j))} \tag{15}$$

Where weights are computed by $w^A(i,j) = [g_A(i,j)]^L$ and $w^B(i,j) = [g_B(i,j)]^L$. L is a constant value in the formulas.

### 3.4 Peng-Wei Wang Metric (PWW)

Wang et al. [44] have discerned the importance of saliency information in the image fusion process, and developed a metric that is based on edge information. However, this metric examines the saliency information on different scales of images. It uses the DWT method to decompose the image into scales.

The final metric is measured with an exponential value $\alpha_i$, this value is used to adjust different relative significance to the different scales.

$$Q_{EP} = \prod_{l=1}^{N} \left(Q_l^{AB/F}\right)^{\alpha_l} \tag{16}$$

### 3.5 Yang Metric

Yang et al. [45] have proposed a quality metric that measures structural similarity. The approach of the metric is the same as in Piella, but a threshold value is used in this metric. With this threshold, it is postulated that the distinction between noise and real information in images is made. This threshold value is taken as 0.75 in the article. The metric's result is decided as follows:

$$\begin{cases} \lambda(w) SSIM(A,F|w) + (1-\lambda(w)) SSIM(B,F|w), & SSIM(A,B|w) \geq 0.75 \\ \max\{SSIM(A,F|w), SSIM(B,F|w)\}, & otherwise \end{cases} \tag{17}$$

### 3.6 Chen Blum Metric

The authors proposed a new metric in 2009 [46]. Image processing applications try to examine images as if they were being processed by the human eye. The human visual system is sensitive to contrast. With this approach, they remark to the contrast values of the images.

## 4 RESULTS AND DISCUSSIONS

### 4.1 Tests

In order to evaluate the success of the proposed method, 17 different methods are compared. Among these methods, relatively state of art ones; MSSF, LP-PCNN, NSCT, CSR, CBF, DWTDE, GF, MWGF, UC, Std, and traditional ones; PCA, Averaging, DCT, DWT, Laplacian Pyramid, Ratio Pyramid are used. These methods are explained in the literature section.

33 image pairs are used to examine these methods. 21 pairs of these images are taken from the Lytro Dataset [47], and the others consisted of commonly used images for fusion studies. In order to make the experiment proper for all methods, the examination is carried out by reducing images to 256x256 size.

Evaluation is performed in two ways; visual and objective. In the visual evaluation, samples of fused images are given, and their error images that are formed by subtracting fused and source images are presented. In objective evaluation, the comparison of methods is performed quantitatively by using six different metrics that can evaluate different characteristics of images.

### 4.2 Visual Comparisons

Two sample test images are used to make visual comparisons. These images have been chosen according to different difficulty level and different prominent features in order to examine the methods from a perspective. Many inferences are made by using fused images and error images that is the subtraction of the first source image from each fused images.

The first sample is Flora Image, it is an easier image than the other. Because it consisting of low frequency and mostly flat areas. Fused images are shown in Figure 5 and error images are displayed in Figure 6. The methods are examined in details separately. In MSSF, the fused image has never close to the correct result and this method produces too many artefacts. GF, DWT, PCA and Averaging methods' error images indicate that these methods create blurred images. Defective pixels appear in almost all of the fused image so these methods underperform compared to others. In Standard Deviation, Unique Color, and DCT methods, blocking-effects appear along block boundaries, and spatial distortions widely occur in their results. In the fused images of CBF and DWTDE methods, local errors have been observed. Error examples are marked with red rectangles in the result images. Although CBF is a gradient-based method, it couldn't able to capture the border of the image. DWTDE, like the CBF method, causes bright shadows and makes distortions at the border of the object. Laplacian Pyramid, Ratio Pyramid, and NSCT methods can capture the outlines, these methods lead to similar unsuccessful results with less qualified images. Incorrect pixels can be seen along with the whole of the objects on the fused images. As can be understood from their error images, the fore object is visible in the form of a silhouette due to failures. The Ratio Pyramid method has less promising results than the other two methods. LP-PCNN method also works incorrectly in the local regions and caused fluctuation, as can be seen in the error image. CSR method can produce well-focused images. But it still cannot be considered to work perfectly, because its fused image has big mistakes under the "Flora" text. The fused images on both CNN, MWGF, and the proposed method are pretty much similar. Composed images that are generated by these methods are successful and outperforms the other methods. In the fused image of CNN method, artefacts are clearly visible on the right edge of the front bottle. It can be said that the proposed method gives a more successfull result.

The second example image is Heart Image, this image is a high frequency and challenging image. It includes so many edges and compelling details. Fused images of all methods are demonstrated in Figure 7 and error images are located in Figure 8. As this image is more challenging for the methods, fused images are more defective. Overall, the evaluations of the fused images are parallel to the previous image comments. As the UC and Std methods cannot work well in sharp-edged regions, the quality of the fused images has dramatically decreased. As in the flora image, MWGF, CNN, and the proposed method yielded the best performance results for this sample image. While the CNN method produces wholly incorrect border of the heart in the fused image. In results of MWGF, it can be clearly seen that it radiates distortion on all sides of the heart. Although the proposed method doesn't produce a more successful fused image than other methods.

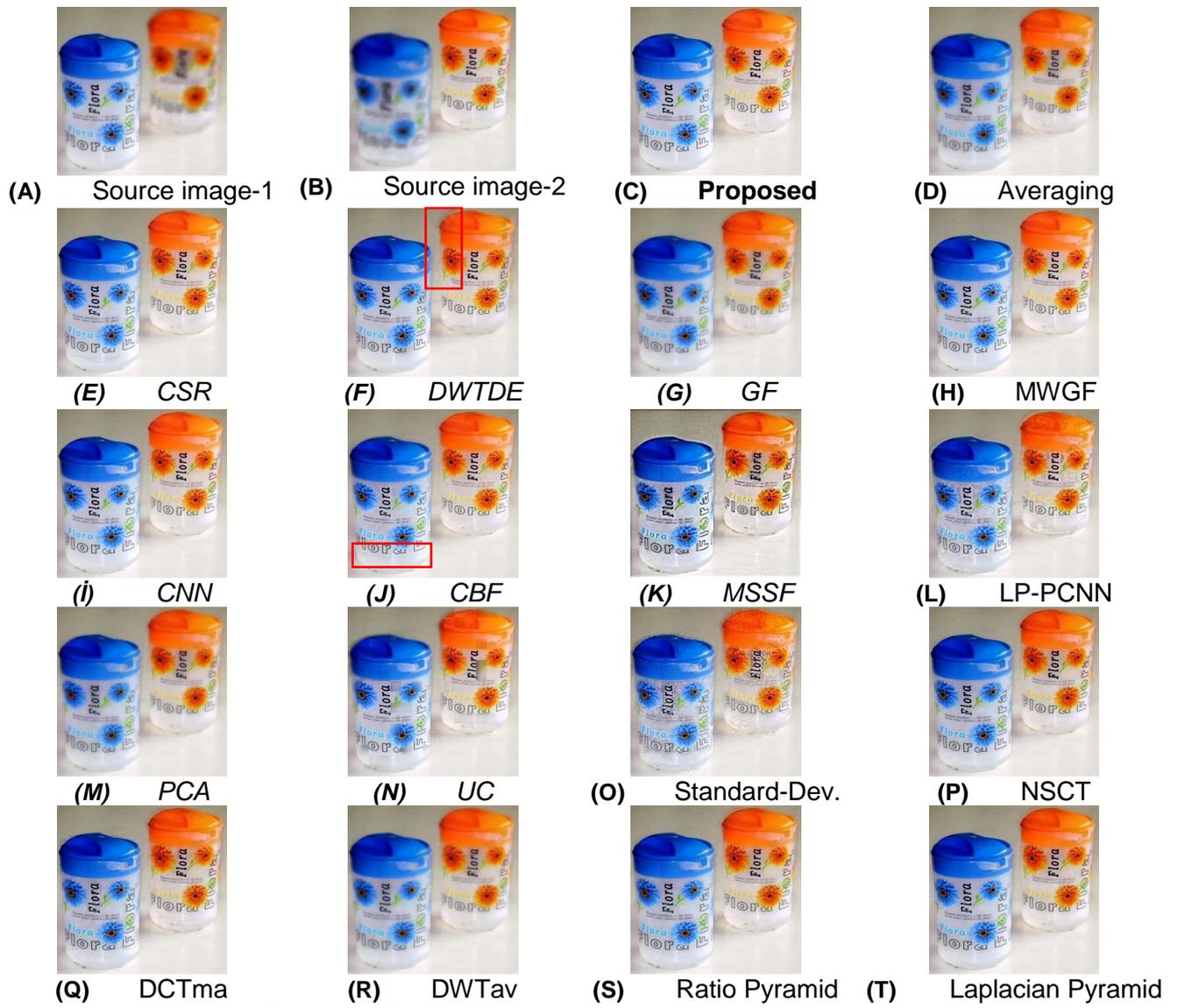

Figure 5 - Fused images of all methods for Flora Image.

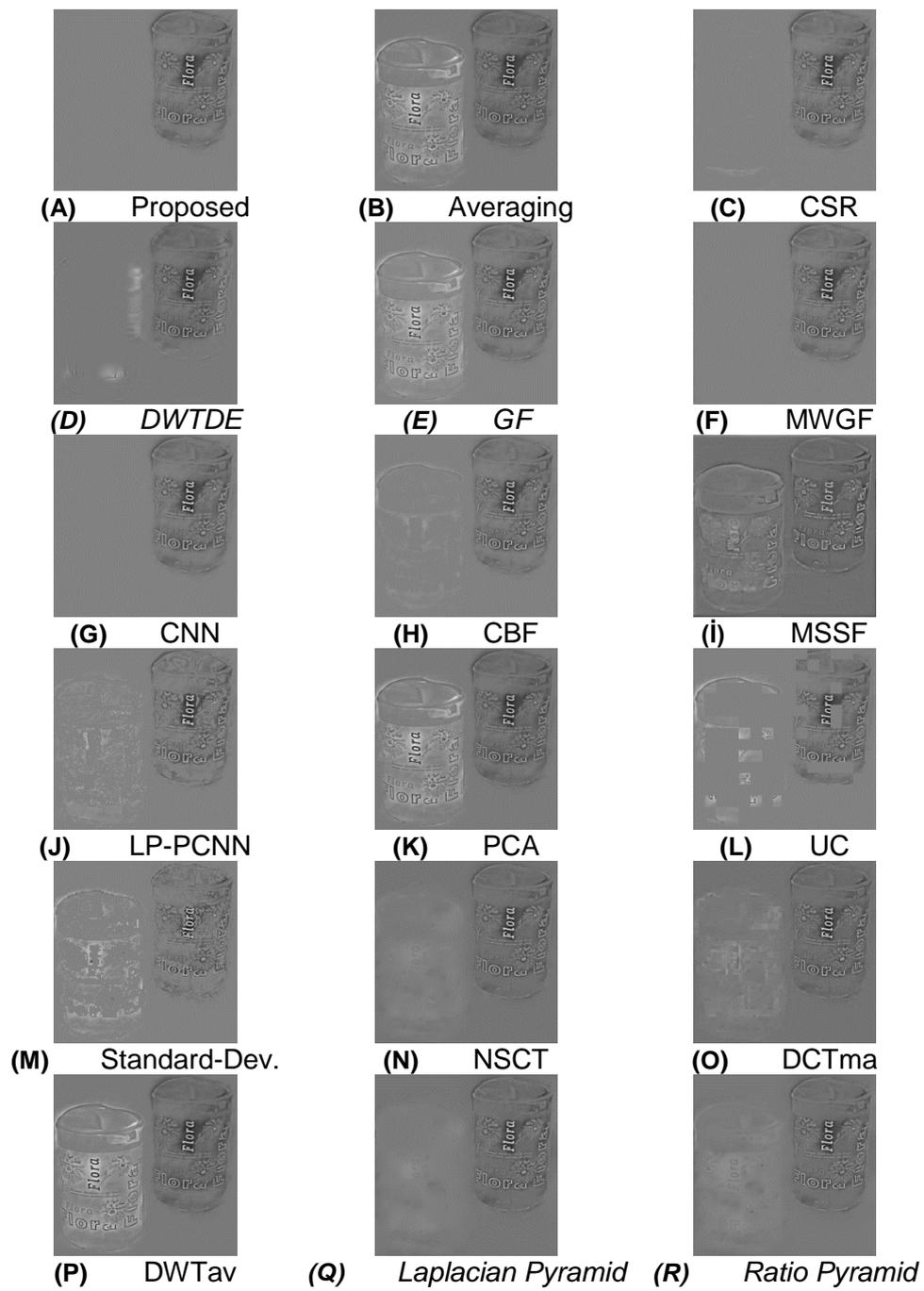

Figure 6 - The error images of all methods for Flora Image.

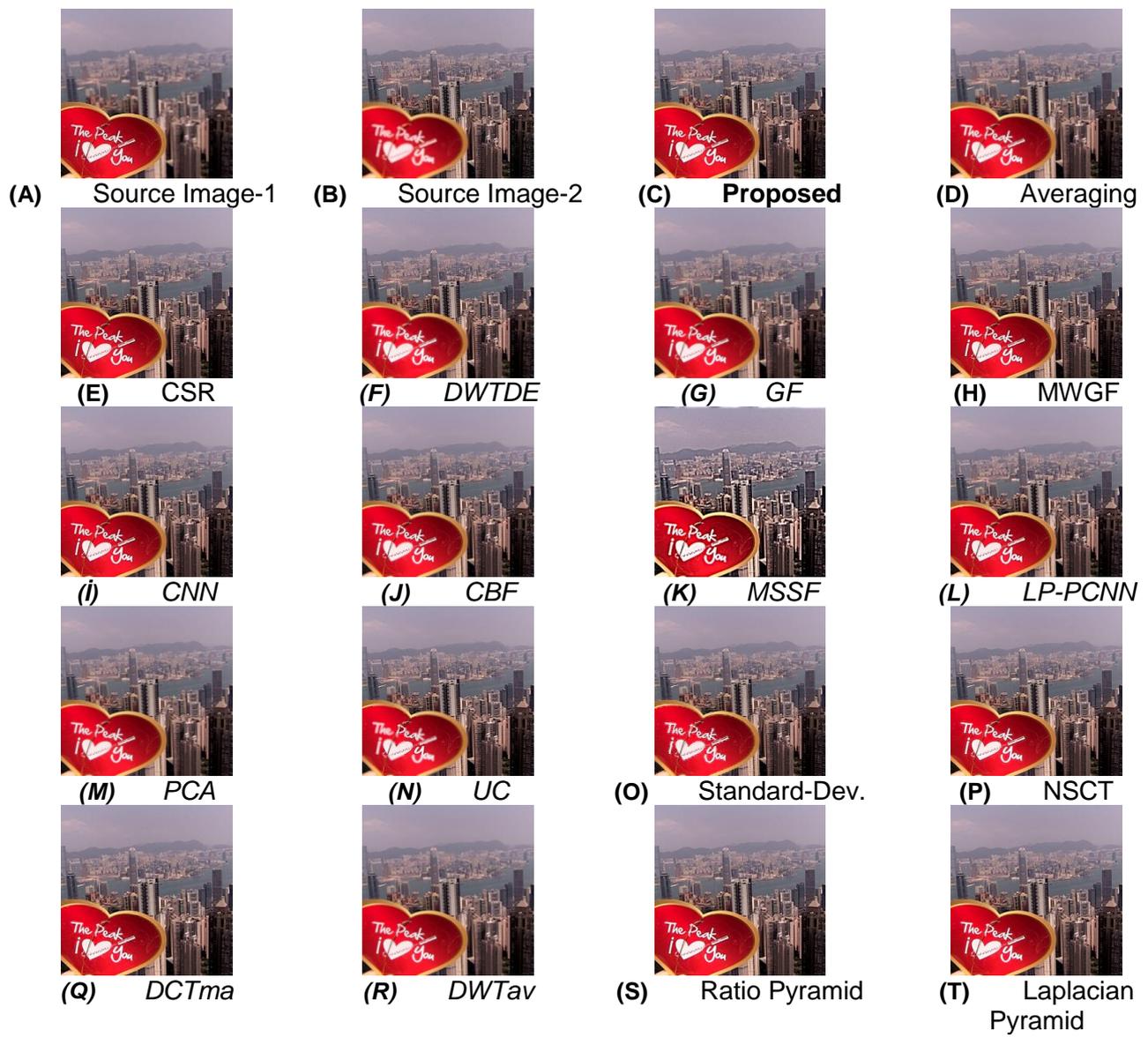

Figure 7 - Fused images of all methods for Heart Image.

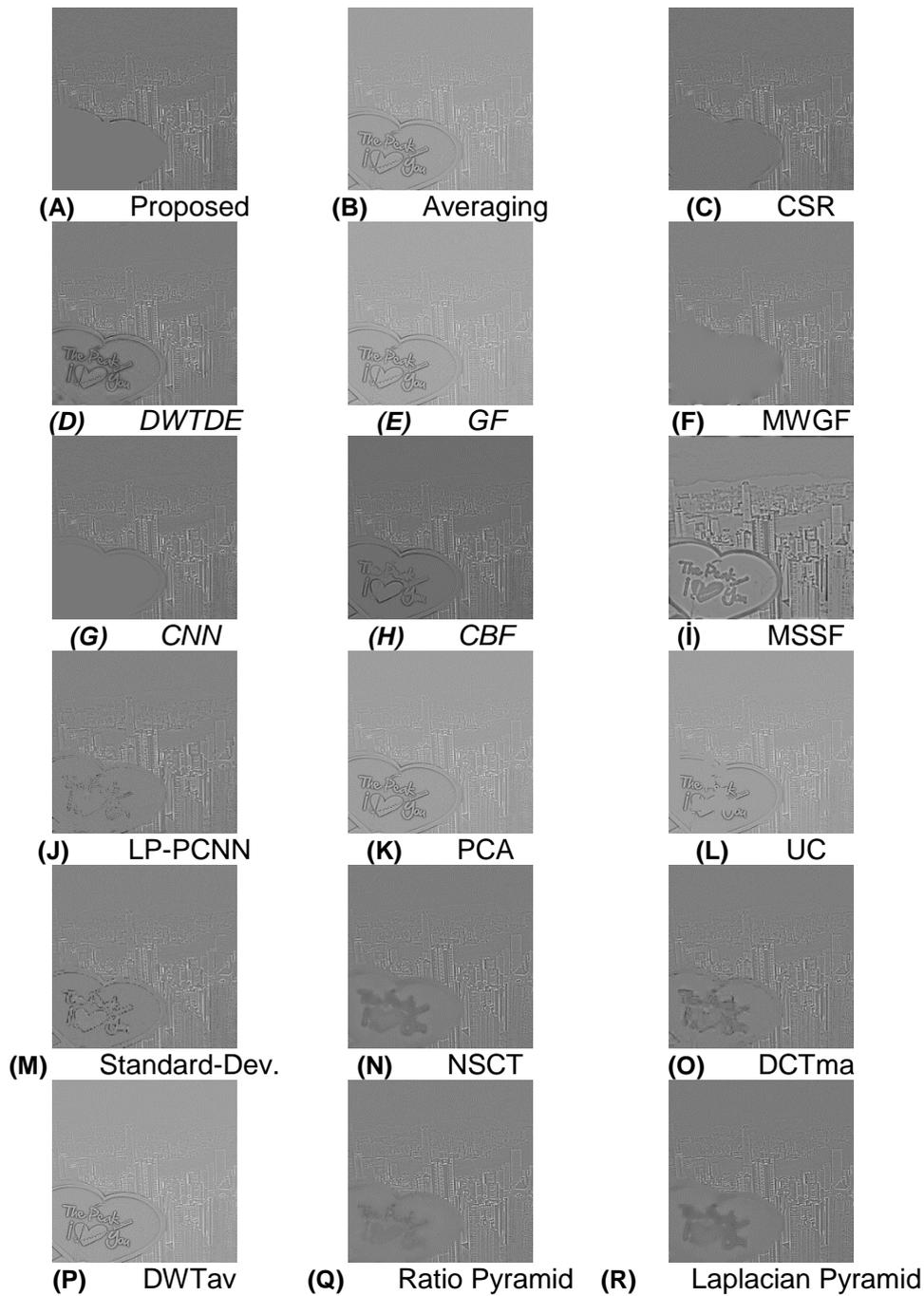

Figure 8 - The error images of all methods for Heart Image.

### 4.3 Objective Comparison

Objective evaluations have been performed by using six well-known metrics; Normalized Mutual Information (NMI), Chen Blum, Peng-Wei Wang Metric (PWW), Xydeas and Petrovic Metric (Qabf), Nonlinear Correlation Information Entropy Metric (NCIE), and Yang Metric. The average results are obtained by taking the results of each image separately and considering the total number of images. These average results are shown in Table 1 and the metrics are described in detail previously. The highest value is shown in bold to indicate the best score and the number of first place among 33 images is given in parentheses near the respective score for each.

According to the objective results; apart from the Chen-Blum metric, the proposed method is superior to the other methods for five metrics. MSSF, DCT, DWT, and GF methods have failed for objective evaluation similar to visual outcomes. MSSF produces the worst results for all metrics. DWT is not successful in motion areas for misregistered images because it is not shift-invariant. GF is not good for capturing the details on the borders of the objects. Averaging, PCA methods have followed these methods and failed to produce satisfying results. NSCT, Pyramid Methods and LP-PCNN methods are not successful overall, but in some metrics, they can produce mid-level success. Metrics measure different characteristic features, so it has revealed that these methods are non-generalized fusion methods. DWTDE, CSR, CBF produce similar results and approach intermediate level success. Surprisingly, although UC and Std methods have poor visual results, they are among the methods that most closely the proposed method in objective evaluation.

Objective evaluations of the proposed method in all metrics (except Chen-Blum) are superior to its competitors. CNN and MWGF have accomplished the closest results to the proposed method. Although the MWGF's success in visual evaluations is close to the proposed method, its objective evaluations are even worse than CNN. CNN's objective evaluations reveal that it is almost as successful as the proposed method in almost every metric. However, CNN's objective evaluation with Chen-Blum metric is slightly better than the proposed method with a small fraction. However, the proposed method in other metrics is superior against CNN and others.

Table 1 - Objective comparison

| Method | MI | NCIE | Qabf | PWW | Yang | Chen-Blum |
| --- | --- | --- | --- | --- | --- | --- |
| **Proposed** | **1.24093**(32) | **0.84967**(27) | **0.72011**(9) | **2.33969**(31) | **0.97765**(12) | 0.80854(13) |
| CSR | 1.01635 | 0.83424 | 0.63535 | 1.10128 | 0.91639 | 0.76680(1) |
| DWTDE | 1.01561 | 0.83528 | 0.65233 | 1.54305 | 0.93395 | 0.76789 |
| GF | 0.91866 | 0.82926 | 0.60938 | 0.35967 | 0.87884 | 0.71480 |
| MWGF | 1.11002 | 0.84033 | 0.69996 | 2.00610 | 0.97426(4) | 0.79482(4) |
| CNN | 1.16617 | 0.84395 | 0.71712(14) | 2.00225 | 0.97679(15) | **0.80886**(14) |
| CBF | 1.02598 | 0.83523 | 0.70320(6) | 0.94365 | 0.95431(2) | 0.77097 |
| MSSF | 0.61250 | 0.81757 | 0.39455 | 0.17136 | 0.75810 | 0.55784 |
| LP-PCNN | 1.19261 | 0.84423 | 0.63080 | 1.05877 | 0.89517 | 0.73963 |
| NSCT | 0.96710 | 0.83247 | 0.68576 | 1.50353 | 0.95181 | 0.77395(1) |
| Averaging | 0.91971 | 0.82932 | 0.61068 | 0.36193 | 0.87861 | 0.71807 |
| PCA | 0.92712 | 0.82974 | 0.61196 | 0.36283 | 0.88270 | 0.71990 |
| UC | 1.21660(1) | 0.84731(6) | 0.71418(4) | 2.10194(1) | 0.96277 | 0.78215 |
| Std | 1.20022 | 0.84555 | 0.66436 | 1.95065(1) | 0.90365 | 0.72939 |
| DCTma | 0.88303 | 0.82811 | 0.61976 | 1.06971 | 0.91395 | 0.72670 |
| DWTav | 0.91706 | 0.82918 | 0.58918 | 0.35834 | 0.86380 | 0.71464 |
| Laplacian | 0.98379 | 0.83338 | 0.69183 | 1.80166 | 0.95187 | 0.77396 |
| Ratio | 0.99130 | 0.83370 | 0.68481 | 1.15869 | 0.94853 | 0.77234 |

## 4.4 Additional Experiments

The contribution of the decision-level fusion step is shown in Table 2. Although it does not contribute much, according to the visual results, it is useful to ensure integrity.

Table 2 - Contribution of decision-level fusion step.

|  | MI | NCIE | Qabf | PWW | Yang | Chen-Blum |
|---|---|---|---|---|---|---|
| w/Decision-Level Fusion | 1.24093 | 0.84967 | 0.72011 | 2.33969 | 0.97765 | 0.80854 |
| wo/Decision-Level Fusion | 1.240497 | 0.84962 | 0.71900 | 2.30735 | 0.97599 | 0.80789 |

Another important experiment is the selection of focus measure function. We aim to highlight the importance of the gradient, so we compare the results using useful and highly preferred gradient-based measure functions. According to the Figure 9, EOG function is suitable as the focus measurement of the proposed method.

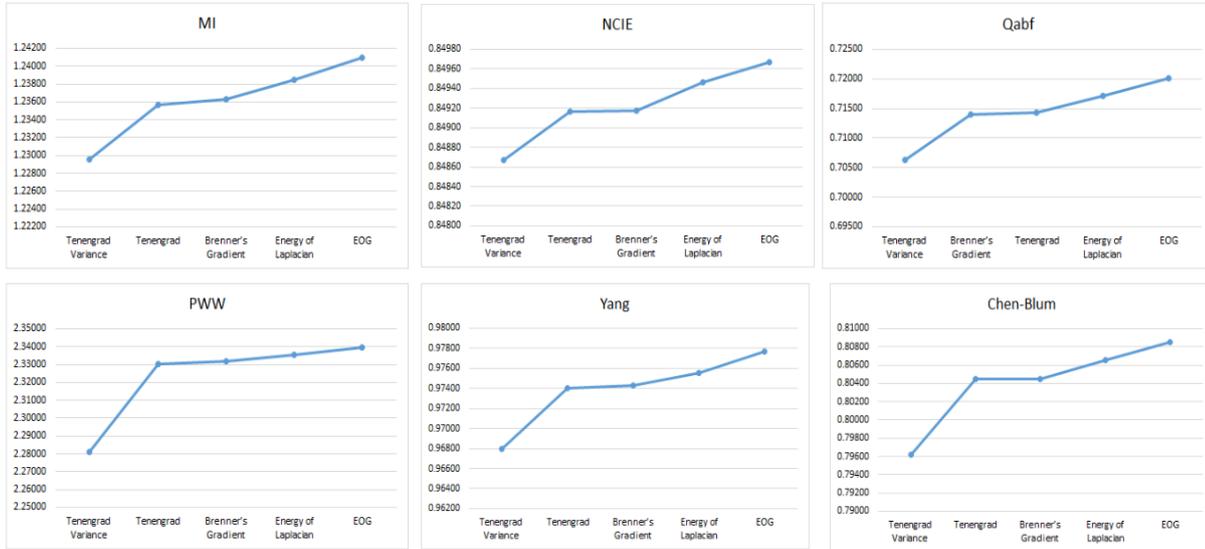

Figure 9 - Comparison of focus measure functions.

Finally, we view the computational efficiency. According to Table 3, It is seen that there are methods that are completed in a shorter time than other methods. The proposed method has a medium run time performance.

Table 3 - Comparison computational efficiency.

| Proposed | CSR | DWTDE | GF | MWGF | CNN |
|---|---|---|---|---|---|
| 4.22 | 29.70 | 1.45 | 4.55 | 0.91 | 31.70 |
| CBF | MSSF | LP-PCNN | NSCT | Averaging | PCA |
| 3.52 | 0.10 | 0.36 | 0.91 | 0.00 | 0.01 |
| UC | Std | DCTma | DWTav | Laplacian | Ratio |
| 0.02 | 1.83 | 0.18 | 0.19 | 0.01 | 0.04 |

## 5 CONCLUSIONS

Completely focused images are necessary for many fields such as medicine, biology, computer science and, etc. Due to the limited depth of field feature of optical lenses, it is difficult to obtain a fully focused image. Researchers suggest multi-focus image fusion methods into this problem. The multi-focus image fusion methods combine the images of the same scene taken at different focal lengths and aim to achieve a fully focused fused image.

The proposed method can capture details, perform well in moving areas. Local histogram equalization uses to enhance the images. The Halftoning-Inverse Halftoning Transform (H-IH Transform) helps to capture the details. The Energy of Gradient and Standard Deviation functions take a part in as focus measures. With the majority voting approach, the fused image has been strengthened. The proposed method is compared with seventeen different methods to prove its performance. Experiments are performed using thirty three pairs of images. The methods are compared visually and objectively.

According to the visual results, it is seen that the proposed method produces near-perfect fused images. CNN and MWGF methods have produced the most similar results visually to the proposed method. UC, Std and DCT methods produced the excessive blocking effect in the fused images. CSR has a medium success in visiual evaluations but it is unable to capture the outlines of the images. MWGF is very close to the proposed method, however it has difficulties to capture the borders of the images.

Six different quantitative metrics are used for objective evaluation. Each metric focuses on different features in the images for evaluations. The conclusions from the objective evaluations are as follows; the proposed method is the most successful method in five of the six metrics. CNN is the most successful method with a slight difference in the Chen-Blum metric. CNN, UC, MWGF, Std methods also presents have prominent results. Although UC and Std methods have simple structures, they have achieved good objective results. CBF has a medium success in objective evaluations too. In CNN, the quality, diversity, and number of the images used during the training process are the factors that affect success. On the contrary, the proposed method is data independent.

In future improvements, we purpose to develop a real-time mobile application to generate a completely focused image.

## DECLARATIONS

**Funding**

Not applicable

**Conflicts of interest/Competing interests (include appropriate disclosures)**

The authors declare that they have no conflicts of interest or no competing financial interests that could have appeared to influence the work reported in this paper.

**Availability of data and material (data transparency)**

The Lytro image dataset can be downloaded from: https://mansournejati.ece.iut.ac.ir/content/lytro-multi-focus-dataset

**Code availability (software application or custom code)**

Codes can be provided by the authors upon request via e-mail.

**Authors' contributions**

SST and MO conceived the study. MO designed and directed the study, SST performed the analyses. SST and MO interpreted the results and drafted the manuscript.